\documentclass[12pt]{amsart}
\usepackage{amssymb,amsmath,amsthm,amsfonts,amsopn,url,color,hyperref,enumerate,mathtools}
\usepackage[normalem]{ulem}
\usepackage{amscd}
\usepackage[normalem]{ulem}
\usepackage[mathscr]{euscript}
\usepackage{mathtools,rotating}
\usepackage{todonotes}
\usepackage{comment}
\usepackage[all,2cell]{xy}
\UseAllTwocells
\input xy
\xyoption{all}
\usepackage{pgf,tikz}
\usepackage{mathrsfs}
\usetikzlibrary{arrows}
\usetikzlibrary{patterns}
\usepackage{tikz,pgfplots}
\usetikzlibrary{matrix, cd}
\usepackage{esvect}
\usepackage{tabularx}
\usepackage{arydshln}
\usepackage{array}

\theoremstyle{definition}
\newtheorem*{question*}{Question}
\theoremstyle{definition}

\author{Hannah Klawa}
\address{Department of Mathematical Sciences \\ Indiana University East \\ Richmond, IN 47374}
\email{hklawa@iu.edu}

\author{Shraddha Rajpal}
\address{Department of Mathematical Sciences \\ Clemson University \\ Clemson, SC 29634}
\email{srajpal@clemson.edu}

\author{Cigole Thomas}
\address{Department of Mathematics \\ Bates College \\ Lewiston, ME 04240}
\email{cthomas5@bates.edu}

\title{GEN AI IN PROOF-BASED MATH COURSES: A PILOT STUDY}

\begin{document}

\begin{abstract}
With the rapid rise of generative AI in higher education, understanding how students use AI is increasingly important. This exploratory study examines student use and
perceptions of generative AI across three proof-based undergraduate mathematics courses:
a first-semester abstract algebra course, a topology course, and a second-semester abstract
algebra course. In each case, course policy permitted some use of generative AI. Drawing on
survey responses and student interviews, we analyze how students engaged with AI tools as well as
their perceptions of generative AI’s usefulness, accuracy and limitations. 
\end{abstract}

\maketitle
\section{Introduction}

As generative AI (GenAI) has become increasingly prevalent, instructors have observed a rise in student submissions that appear to reflect the use of these tools but GenAI detection tools have not proven reliable \cite{Originality, Giray2024, LIANG2023, BypassgenAI}. While GenAI is likely affecting students’ learning experiences, there were relatively few studies available on student perceptions of integrating AI into proof-based mathematics courses at the time of study although there are increasingly more studies being done in this area \cite{Amel2025, Wei2024,Stylianides2024, Yoon2026, Yoon2024}.
Understanding students' use and perceptions of gen AI may help educators design effective learning experiences \cite{Chan2024}.

This was a small exploratory pilot study that was designed to provide preliminary insight into students' use and perceptions of GenAI in upper-level proof-based mathematics courses. The study reflects on themes and patterns that emerged from surveys and interviews with students enrolled in three proof-based courses during the spring of 2025: a first-semester undergraduate abstract algebra course, an undergraduate topology course, and a second-semester undergraduate abstract algebra course. In these courses, a policy was implemented that allowed students to use GenAI with some constraints and minimal guidance. 

In addition to examining student use of GenAI, this study shares insights from student perspectives. We focus on how students used GenAI, how they perceived its usefulness and limitations, and what their responses suggest about accuracy, responsibility, engagement, and proof-writing in advanced mathematics courses.
In particular, we consider the following questions: 
\begin{enumerate}
    \item How do students use gen AI in the course?
    \item What are students' perceptions of gen AI's helpfulness to their learning and course experience?
    \item What are the students' perceptions on the impact of gen AI on student engagement with peers and instructors? 
    \item How do students view gen AI's reliability and accuracy for proof-based problems? 
\end{enumerate}
The paper begins with an overview of related research and the study's methods. We then present and discuss the survey and interview results in relation to existing research and future considerations.

\section{Related Research}
The use of GenAI has become increasingly prevalent in math education \cite{FSPJ2025,GARCIA2025,diffeqchatgpt,Park2024,Updatingcalc,Walkington26,Calculuschatgpt}. In addition to commonly used GenAI platforms, there are AI tools geared towards tutoring such as LeanTutor, GPT-tutor, and others \cite{chen2026,Nikolic26,LeanTutor, VSMP2024, ZSP25}. 
\subsection{Use and Accuracy of AI}
There are many studies looking at student perceptions of AI use in higher education \cite{AlAbdullatif2024,Gomez2026,Chan2024,GARCIA2025,Johnston2024,Kim2025,Yeung2026}.
Uses of GenAI include using it for searching, conceptual understanding, feedback and revising proofs \cite{LLMproofsingeo, Park2024, Yoon2026}.
While many students in higher education are utilizing GenAI, there are still a significant percentage of students who have never used gen AI platforms and/or have ethical concerns about its usage  \cite{Gomez2026, Chan2024, Chongchitnan2025, Feeney2026-2, Eddine2025}. Although GenAI continues to improve and has even been used to prove new results in  mathematics research, current AI systems still demonstrate inconsistent performance in mathematical reasoning \cite{firstproofsecondbatch, LLMmathreasoning,mathcapabilitychatgpt}.
While students find gen AI helpful, they have mixed feelings about the accuracy of AI \cite{Collins2024, Gomez2026,Chan2024, Yoon2024}.

Some researchers have  proposed a frame-work called Students Interactive Proving Experience with AI (SIPE-AI) to describe how students interact with AI-generated proof attempts and how their use is shaped by their conceptions of proof, conceptions of AI, and ethical considerations \cite{Yoon2024}. Others have explored the student use of and interaction with gen AI using this framework \cite{Amel2025}. 

\subsection{Helpfulness of Gen AI}
Students' perceptions of AI in mathematics are not uniform, but may vary according to prior experience, academic context, and patterns of use \cite{Ergene2025}.
Students often perceive GenAI as helpful in understanding mathematical concepts and for evaluating and revising mathematical arguments through feedback \cite{mopreAIandgrouptheory, Park2024, Sassanapitax2026, Yoon2026}. Use of GenAI in a structured way can help achieve higher learning outcomes \cite{Pallant2026}. However, there are concerns about its consistent good performance, especially when it comes to advanced mathematics \cite{Collins2024, mathcapabilitychatgpt}.  This suggests that AI can be a helpful tool, but its effectiveness depends on students’ ability to critically evaluate AI-generated reasoning and maintain their own mathematical engagement \cite{Yoon2024}.

\subsection{Impact of Gen AI on Student Engagement}
Peer-interactions and instructor-student interactions plays an important role in creating a quality learning experience \cite{Garrison1999, Wood1994, You2017, Zhang2022}.  While gen AI can be used to help improve communication and streamline group teamwork, it can also reduce the need for communication with instructors and peers~\cite{Chang2026, Juan2026, Wei2026}. In this paper, we explore student perceptions both of genAI's impact on peer interactions and engagement with the instructor.

\section{Methods}

\subsection{Description of Program and Courses}
\subsubsection{The Undergraduate Mathematics Program and Departmental Context}
Indiana University East is a small regional commuter campus that offers a combination of online and on-campus courses and programs. Over the past three years, total student enrollment has remained close to 3,000.  During the fall of 2024 semester, the B.S. in Mathematics program enrolled 268 students, with 190 enrolled part-time and 210 classified as non-traditional in age. Notably, the mathematics degree can be completed entirely online.

Upper-level, proof-based courses in the program are taken both by IU East students and by students intending to transfer credits to other institutions. The three courses examined in this study were offered in an asynchronous online format.

As of spring of 2025, the Department of Mathematics consisted of 3 tenure-track faculty, 4 lecturers, 1 full-time visiting faculty member, and 12 part-time faculty. For upper-level proof-based courses, the average enrollment over the past year was approximately 19 students per section.

\subsubsection{The Courses Involved in the Study}
This study focuses on three upper-level proof-based mathematics in the spring of 2025. 
All three courses were delivered in an asynchronous online format and shared a consistent structure that combined textbooks with instructor-generated videos for instructional content. Assessment methods included graded online discussions, quizzes, and weekly written assignments submitted in two stages. Students first submitted their work and received feedback and an initial grade, then revised their submissions based on that feedback for final grading. 

The first course, MATH-M403 - Introduction to Modern Algebra I (a first semester undergraduate abstract algebra course), had 41 students enrolled, 8 of whom withdrew. Students taking this course are expected to have previously completed an introduction to proof-writing course. The assessments for this course included a midterm and a final exam, in addition to the weekly submitted assignments. The second course, MATH-M404 - Introduction to Modern Algebra II (a second semester of undergraduate abstract algebra), had 12 students enrolled. This course was taught in a combined section with a graduate-level algebra topics course designed for students seeking credentials to teach dual-credit mathematics at the high school or community college level. In addition to the weekly submitted assignments, there was a midsemester group project and a final group project. The third course, MATH-M421 - Introduction to Topology, had 14 students enrolled, 2 of whom withdrew. This course also had a midsemester group project and final group project in addition to the weekly submitted assignments. Together, these three courses provide a useful setting for exploring students' use and perceptions of GenAI use in mathematical reasoning due to their emphasis on abstract concepts and proof writing.

\subsubsection{The Gen AI Policy Implemented}
During the spring of 2025, the following syllabus statement was used to outline the policy on GenAI in each course. At the time, the only GenAI officially approved for use with university-internal data at Indiana University East was Microsoft CoPilot. This approval was based on its compliance with university data security guidelines; it was authorized for use with university-internal data and did not utilize input from IU accounts for training purposes. Some language in the policy below was adapted from (\cite{Hodgson}) and Indiana University’s GenAI policy at the time of implementation (\cite{IU}).
\begin{quote}
    The use of generative AI (GenAI) is permitted in this course for all assignments and assessments. When using generative AI, extra caution must be taken to protect personal data and the data of others and respect copyrights. Due to this, we will be utilizing Microsoft CoPilot through our IU accounts which has been approved for university use because it keeps data within the IU/Copilot system.
    \begin{itemize}
\item For homework, you will use Microsoft CoPilot through your IU account which has been approved for university use. After using it, you will need to evaluate the response and provide appropriate justifications if the resulting solution or proof is correct and has sufficient work shown. In your submitted work, you must only utilize previously learned results in the course and must give reference results by number from the textbook or lecture notes. 
\item For discussion assignments, you must state which parts are generated by GAI unless you have only used it for grammar and spelling revisions. In discussion, you are encouraged to include results obtained via GAI but you should critically evaluate these yourself and enlist the help of your peers in evaluating.
\end{itemize}
 However, students should note that all large language models (basis of GAI) still have a large tendency to make up incorrect facts or fake citations, produce inaccurate outputs, or generate highly offensive products. You will be responsible for any inaccurate, biased, offensive, or otherwise unethical content you submit regardless of whether it originally comes from you or a GAI platform. You must take personal responsibility for your submitted work.   

\end{quote}
For each course, the homepage in Canvas also included a welcome message with the following statement. 
\begin{quote} Welcome to MATH-M4XX - Course Name. Please take time to navigate to the syllabus in the left navigation bar. All the learning material will be within the modules. As you will see in the syllabus, you may utilize and are encouraged to utilize Microsoft Copilot using your IU account [Link included to Indiana University's webpage on using Microsoft 365 Copilot through your IU account] for first draft of homework. Please be very cautious!!! You take personal responsibility for all submissions regardless of whether the text was generated by Copilot. We will critically evaluate any responses obtained through Copilot and must justify all work using theorems/results covered up to the point at which we are in the course. Within mathematics, there may be different ways to prove the same thing so what you obtain may not be appropriate for submission and often you might obtain a totally nonsensical answer.

If you have any questions at all, please let me know. I am looking forward to working with all of you this semester!
\end{quote}
\subsection{Method of Data Collection}
Based on the instructor's experience from teaching the courses and the authors' interests, five key themes were identified: ease of use, helpfulness to students, students' perception of GenAI in math courses, students' perception of dependence on GenAI and perception of classmates' usage of GenAI. 

The survey and the zoom interview questions  designed based on these themes were used for data collection. 
The data were collected through voluntary surveys and zoom interviews conducted at the conclusion of the semester. Before distributing surveys or conducting interviews, Institutional Review Board (IRB) exempt status was obtained both from Indiana University and Clemson University. To mitigate any potential pressure on students to participate, recruitment emails were sent out by the Dean of Natural Science and Mathematics at IU East. Additionally, the initial survey recruitment email was sent after the courses had ended further ensuring that participation could not impact students' grades. Survey and interview data were collected by two non-instructor researchers, who anonymized the responses before involving the instructor researcher in the analysis. Participation was entirely voluntary, and students were informed that their responses would remain confidential and de-identified. All data were stored securely in accordance with institutional research ethics protocols. These steps were taken to address potential ethical concerns that might arise with student participation in this study (\cite{LL2013}).

The survey was administered using Qualtrics, a secure, university-supported online survey platform. It consisted of 26 questions including a mix of open-ended questions and likert scaled questions. A total of 59 students were invited to participate including those enrolled in a graduate algebra topics course.  However, because no responses were received from the graduate course, results reported here focus only on the three undergraduate proof-based courses. Specifically, there were a total of 10 survey responses from students in MATH-M403 - Introduction to Modern Algebra I, 4 survey responses from students in MATH-M404 - Introduction to Modern Algebra II and 5 responses from students in MATH-M421 - Introduction to Topology I. It is important to note that two students completed surveys for more than one course (one for both M403 and M421, and another for both M404 and M421). In total, the study included 19 survey responses from 17 students resulting in an overall response rate of approximately $29\%$.

In addition to the surveys, 4 students participated in the interview phase of the study. Interviews were conducted via Zoom and lasted approximately 15–20 minutes each and followed a semi-structured protocol. Interviewers began each session by introducing the purpose of the interview and obtaining consent to record the session. The interviews aimed to capture a deeper understanding of students’ experiences, attitudes, and perspectives regarding the use of GenAI in proof-based math courses. After anonymizing, the open-ended responses within the Qualtrics survey and the interview transcripts were coded by all three authors and used in presenting the results in the following section. 
\section{Results and Discussion}

\subsection{How Students Utilized AI in the Course}
In this study, 8 out of 19 participants reported they did not use GenAI for their coursework although using it was allowed by course policy. Notably, 7 of the participants viewed utilizing GenAI for coursework as cheating.

All of the respondents who utilized AI for the course also reported asking follow-up questions. 
Some people who reported using GenAI asked AI to solve questions directly while others used it for brainstorming, providing feedback, explaining concepts, and references for further learning. 
Most students using GenAI found it most beneficial for finding information rather than actually proving statements. For instance, one student noted: 
\begin{quote}
    It did help me go deeper into some of the history and the concepts. That’s really where I found the greatest use.
\end{quote}
Another noted: 

\begin{quote}
     I ended up using it as a complex dictionary to explain concepts more deeply until I could really understand them. It also helped me with concepts that I needed to rely on from earlier classes.
\end{quote}
Some also used it for checking proofs or finding starting points: 
\begin{quote}
    Some questions I gave it the whole proof to see what it would give as a starting point and to see if it seemed correct. Sometimes I asked for definitions or pictures. But most often I used it to recall things that took much longer to try and find in my notes just to refresh my memory and help me give better reasoning.
\end{quote}

\subsection{Helpfulness to the Student Learning and Course Experience}
\begin{figure}
\begin{center}
    \includegraphics[scale=0.5]{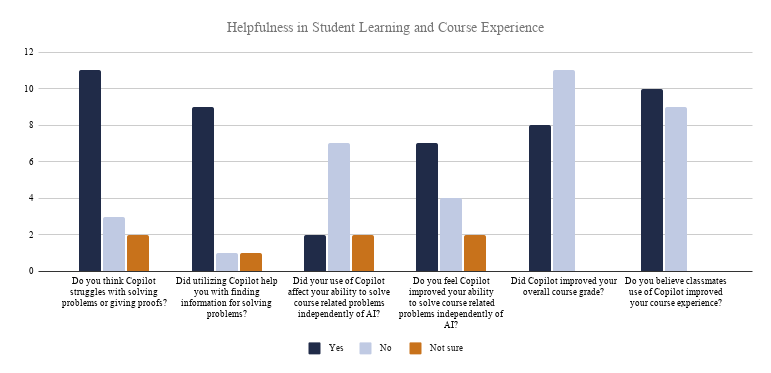}
   \caption{}
    \label{fig:help}
    \end{center}
    \end{figure}

The survey data suggests that students generally found MS Copilot to be a helpful tool for learning in a proof-based mathematics course (see Figure \ref{fig:help}).

\begin{figure}
\begin{center}

   \includegraphics[scale=0.9]{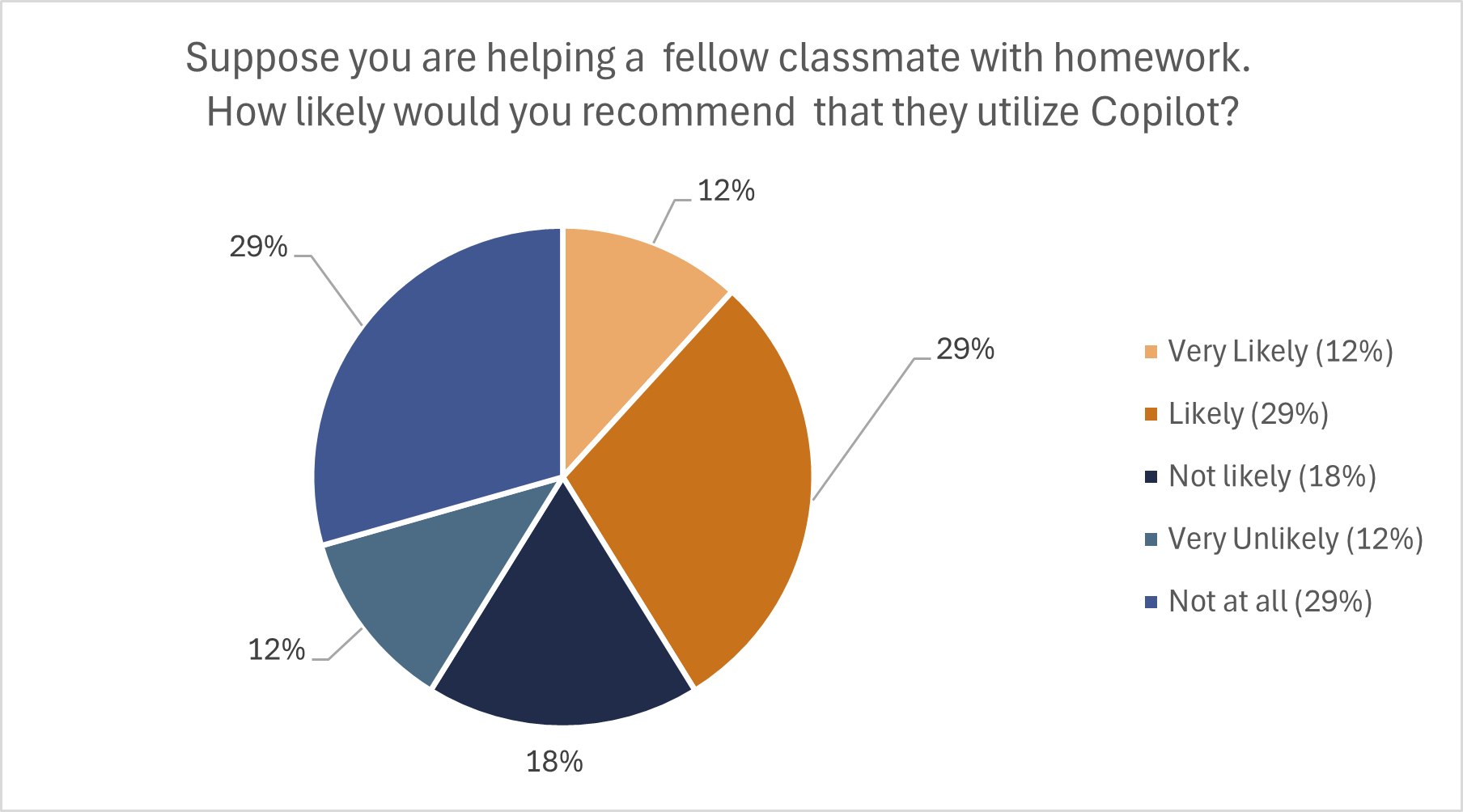} 
   \caption{}
   \label{fig:classmate}
\end{center}
\end{figure}

 From an examination of all responses in Figure~\ref{fig:help}, it seems possible that students interpreted the phrase ``affect your ability to solve course related problems independently of AI'' as a negative impact. A majority (11 out of 19) of respondents appreciated its inclusion in the course, particularly noting its usefulness in finding information and assisting with problem-solving. However, about half of students (9 out of 19) did not support the idea of incorporating GenAI across all mathematics courses, indicating a preference for selective and context-dependent use. While several participants reported that Copilot helped them understand material or brainstorm approaches to problems, its impact on independent problem-solving was more mixed, with responses distributed across “yes,” “no,” and “not sure.” There were also mixed feelings about whether they would recommend utilizing MS Copilot to a fellow student for homework help (Figure \ref{fig:classmate}).

 Notably, many students acknowledged that Copilot struggles with solving complex problems or providing complete proofs, highlighting limitations in GenAI’s capacity at the time of data collection to engage with advanced mathematical reasoning. These mixed feelings are evident from the student comments from open ended questions in the survey and interviews.

One student while recognizing that the responses were not always accurate was particularly enthusiastic about the helpfulness of AI in understanding the material as evident from the following quote. 
\begin{quote}
It was a godsend, honestly.  It was a very effective way to work through homework problems especially, and I feel like I ended up understanding the material more than if it wasn’t used. You could ask it to expand specific parts of answers if something wasn’t clear, and the fact it wasn’t always right was really helpful as it made you think through the problem, not just accept the answer. 
\end{quote}

One student noted that finding the errors was like ``debugging code" as opposed to the usual proof writing process. 
\begin{quote}
It [AI] seems like it was generally pretty good. But it would, it would just make different kinds of mistakes. Like that a human wouldn't. That were very obvious to spot, and so we would point those out in the discussion, and then, presumably, the student would run it through the AI a second time, and then we'd get new issues. And it was, it was a different experience than having an actual conversation with someone and trying to work through a problem. .... \\

 It [Using AI] was more like debugging code than writing a proof.

\end{quote}

Some students found it helpful with writing aspects related to the course such as grammar and wording. For instance, one student said: 
\begin{quote}

Copilot still has a long way to go in proof-writing, but it does an okay job verifying accuracy of proofs. Because it gives short answers, I often had to ask many follow-on questions in order to better understand the specific topics. Also, if I knew how I wanted to respond in a discussion post, I would ask Copilot to help me write the post because sometimes I stumbled with my words.

\end{quote}

A few students raised concerns that GenAI, ``takes away the pride of ownership" and ``compromises the learning process'' as noted in the following quotes. 
\begin{quote}
    
    It takes away the pride of ownership. Especially with writing, which is key in my work. If you rely too much on any tool, it takes away from your voice and expertise.
\end{quote}
\begin{quote}
    I worry that especially in pure math, the value is lost for those students who rely on GenAI to help them solve hard problems.
\end{quote}
\begin{quote}
I’m here primarily to learn, so I avoid tools that might compromise the learning process.

\end{quote}

It is important to note that students' perception of deeper comprehension could be different from reality.  Students noted that using AI for direct solution generation or discussion posts sometimes detracted from their sense of personal achievement and mastery.  On the other hand, some felt that AI facilitated critical thinking through error detection and “debugging” exercises, where students evaluated AI-generated proofs or reasoning and identified inaccuracies, effectively learning proof-writing skills by critiquing flawed examples. A key concern raised in both student comments and the literature is that utilizing GenAI may decrease students' critical thinking and ability to independently generate proofs \cite{ON2024}. While proof validation is a necessary skill for a mathematician to have, it is one that is not necessarily explicitly taught in the undergraduate mathematics curriculum \cite{Selden}. Based on our current study, it seems that the exercise of validating AI generated proofs may provide a way of increasing students' critical thinking and understanding of proofs, although more research on this particular aspect is needed and validating proofs is only one aspect of mathematical skills \cite{Sriraman2026, Urhan2026, Weatherall2026}. 

\subsection{Impact on Student Engagement}

Encouragingly, students did not generally report that using Copilot reduced their engagement with instructors. 
\begin{figure}
\begin{center}
\includegraphics[scale=0.5]{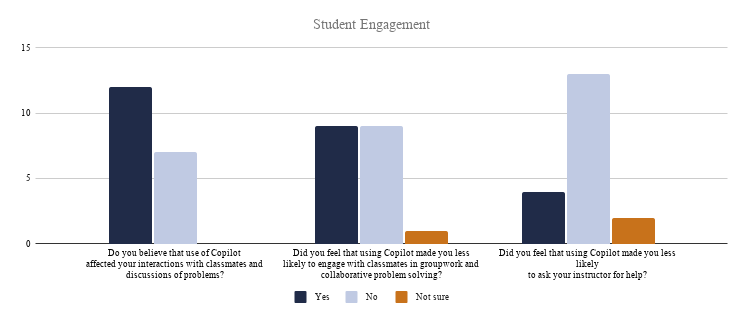}
\caption{}
\label{fig:engagement}
\end{center}
\end{figure} 
On the contrary, many continued to view direct instructor support as a central part of their learning process.

However, the participants were split on whether it affected their interactions with their classmates (Figure \ref{fig:engagement}).
Findings related to student engagement suggest that the use of MS Copilot influenced peer interaction in nuanced ways. While some students noted changes in how they engaged with classmates or discussed course material, these shifts did not appear to undermine collaborative problem solving in a consistent or widespread manner. Students’ perceptions of their groupwork experiences varied.  
 
A couple of students thought in general that the use of AI improved the level of discussion. However, they felt that some students were posting responses ``entirely AI generated''. 
\begin{quote}
    Generally I feel the average level of discussion and accuracy of examples by students was higher. On the other hand there were some replies that contained very little interesting information and were probably entirely AI generated.
\end{quote}
\begin{quote}
    I could clearly tell when other students just copy/pasted what copilot said and that was detrimental because I knew I couldn't really interact with them. In other cases, using Copilot highlighted missing gaps in knowledge and sparked curiosity in related topics that elevated the conversations. 
\end{quote}
Another student expressed their frustation in feeling like they are talking to ``unfeeling robots''. 
\begin{quote}
    Some of the other students in discussion were clearly using Copilot to generate discussion responses, and that felt disingenuous. I’m here to have a discussion with humans, not to have my questions run through an unfeeling robot.
\end{quote}

Overall, there seemed to be mixed feelings about how GenAI impacts peer interactions. But students did not seem to feel an impact on their instructor interactions. This agrees with other findings that the use of GenAI has the potential to be used to enhance peer-collaboration with carefully structured use \cite{Juan2026}.

\subsection{Students' Perception of Reliability and Accuracy}
\begin{figure}
\begin{center}
\includegraphics[scale=0.6]{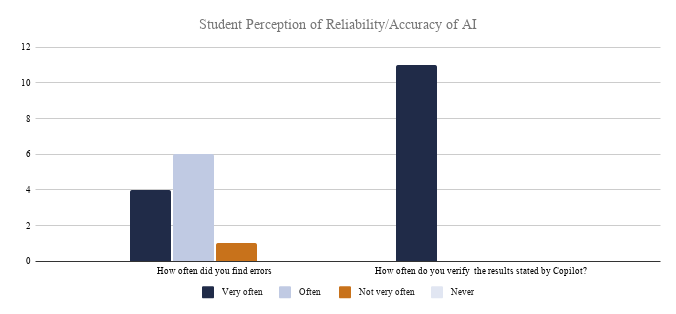} 
\caption{}
\label{fig:reliability}
\end{center}
\end{figure}

Students’ perceptions of MS Copilot’s reliability reveal a cautious and discerning approach to AI-generated content (Figure \ref{fig:reliability}). 
Many students regularly identified inaccuracies in the tool’s responses and were consistently vigilant in verifying the information it provided. However, it is worth noting that verifying AI outputs was part of the course policy which may have influenced the student response to this question. 
 Rather than relying on Copilot uncritically, students treated it as a supplementary aid whose output required careful evaluation. 

One student considered AI useful for collecting information but the information can be inaccurate sometimes. 
\begin{quote}
    AI can be a highly efficient tool for collecting information. However, ChatGPT sometimes presents inaccurate information with a confident tone.
\end{quote}
Another student felt that Copilot is ``wildly inaccurate'' for advanced proof-based courses.
\begin{quote}
    Copilot can’t help a ton with the super difficult questions or problems... it can be wildly inaccurate when it comes to proofs and abstract concepts.
\end{quote}
Generally, students recognized the need for critically evaluating AI responses.
\subsection{Ease of Use}\label{Fig_ease}
\begin{figure}
\begin{center}
\includegraphics[scale=0.8]{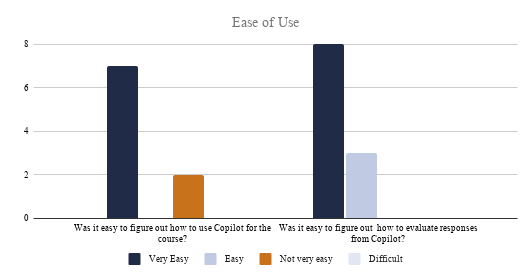}
\caption{}
\label{fig:ease}
\end{center}
\end{figure}

Overall, students found MS Copilot intuitive to navigate and apply within the course context (Figure \ref{fig:ease}). 

 Similarly, students found it straightforward to evaluate the responses provided by Copilot.   Three students mentioned  that some sort of introductory tutorial or workshop would have been helpful with one of them mentioning that: 
\begin{quote}
     I think it'd be good to sort of give an overview on use of the AI... like a five-minute demonstration video or something, or like example outputs... just anything to push students towards at least trying it.
\end{quote}

Of note, because the policy allowing the use of AI required a more rigorous system of references to textbook numbering of theorems and page numbers for results and definitions compared to what might be the typical expectations for proofs, this added ``tedium'' as noted by a student:
\begin{quote}
    We had to cite specific theorems in the textbook by like number... which hasn’t been a thing that I’ve had to do in previous classes... it just adds a lot of tedium.
\end{quote}
\section{Limitations}
 It is
important to acknowledge the limitations of this study such as a small participant pool.
Student responses may have been influenced by the course policy requiring verification of
AI outputs, potentially amplifying caution in perceptions of reliability. Furthermore, the
study focused on limited number of topics in advanced mathematics (topology and abstract
algebra) and a specific AI tool (MS Copilot) which may limit applicability to other courses,
disciplines, or AI platforms. In addition, the fact that the surveys and the interviews were
not anonymous may have impacted some of the student responses. Another factor to note
is the fact that a higher percentage of interviewees $(50\%)$ didn’t use Copilot in comparison
to the survey participants but it is not clear whether this represents the actual percentage
of AI usage among the students in the courses. Finally, self-reported data on AI use and
perceptions could be affected by social desirability or recall biases, especially given that
some students viewed AI use as potentially “cheating” \cite{Yao2026}.  Despite these limitations, the study
provides exploratory insights into the opportunities and challenges of integrating GenAI into advanced mathematics education.
\section{Conclusion and Future Considerations}
Given limited instructions on how to use GenAI, students utilized it in a wide variety of ways or did not utilize it at all. The majority found it to be helpful in learning the course material and improving the course experience. While a majority thought it improved their problem solving ability independently, more than half did not feel like it improved their course grade. While the use of MS Copilot does not seem to have affected the student interaction with instructors, there seem to be varying opinion on how it affected student-to-student interaction. Some students thought the use of GenAI led to deeper student discussions whereas some felt a loss of human connection due to suspicions of fully AI generated discussion posts. In general, students were skeptical of AI responses and verified all AI generated results (which was a course requirement). Most of the students found MS Copilot easy to use while a few students could have benefited from a short tutorial in the context of the course material and policies.  Overall these findings suggest that students have differing perceptions regarding GenAI but there is potential for it to be used in a constructive way if used with proper guidance.

A few students pointed out that it might be useful to have an instructor demonstration on utilizing GenAI. This agrees with one of the practical implications listed in \cite{GARCIA2025} where they mention that ``Ease of use, facilitating conditions, and performance expectancy may
also be enhanced by developers and instructors alike; wherein, the application and interface may be demonstrated among students to teach them better actual utilization of the different tools.''  It may be helpful for students to see not only how to phrase effective prompts, but also how to critically evaluate AI outputs, find errors in the logic, find accurate supporting references to verify results  using established mathematical results and document the references with proper citations.

Another concern is that utilizing GenAI may decrease students' critical thinking and ability to independently generate proofs \cite{ON2024}. To address this, further research is needed on creating and refining assessments within proof-based courses that promote critical thinking.
Looking ahead, future research with larger and more diverse cohorts could help identify best practices for integrating AI into proof-based mathematics instruction in ways that both harness its strengths while preserving students' abilities to reason independently.

\section*{Acknowledgments}
We would like to thank Markus Pomper the Dean of Natural Science and Mathematics at Indiana University East for his help with sending out emails to advertise the survey to students. We also thank Justin Hodgson and the Digital Gardner Initiative at Indiana University for encouraging the incorporation of GenAI into courses and for policy guidance which helped motivate this study. The first named author is grateful for the Summer Faculty Fellowship at Indiana University East which provided support during part of the study.

\bibliographystyle{plain}
\bibliography{bibliography}

\end{document}